\pdfoutput=1

\documentclass[11pt]{article}
\usepackage{acl}

\usepackage{times}
\usepackage{latexsym}
\usepackage{amsmath}
\usepackage{amssymb}
\usepackage{multirow}
\usepackage{booktabs}
\usepackage{subcaption}
\usepackage[T1]{fontenc}
\usepackage[utf8]{inputenc}
\usepackage{microtype}
\usepackage{inconsolata}
\usepackage{graphicx}
\usepackage{xcolor}
\usepackage{float}
\usepackage{enumitem}
\usepackage{colortbl}
\usepackage{soul}
\usepackage{tcolorbox}
\usepackage[table]{xcolor}

\newtcolorbox{promptbox}{
  colback=yellow!6,        
  colframe=black!55,       
  boxrule=0.5pt,           
  arc=1pt,                 
  left=1pt,
  right=1pt,
  top=1pt,
  bottom=1pt,
  fontupper=\ttfamily\footnotesize,
}

\title{Attention-Steered Vision-Language Models for Sign Language Translation}

\author{Meibo Hu$^{1}$, Guohao Sun$^{1}$, Annemarie D. Ross$^{3}$, Sheng Li$^{2}$, Zhiqiang Tao$^{1}$ \\
  $^{1}$Rochester Institute of Technology, $^{2}$University of Virginia \\
  $^{3}$National Technical Institute for the Deaf / Rochester Institute of Technology \\
}

\begin{document}
\maketitle

\begin{abstract}

Vision-language models (VLMs) have emerged as a powerful framework for multimodal video understanding. However, they remain limited in the sign language translation task, where we identify a key failure mode of existing VLM-based translators: \emph{poor spatial-temporal visual grounding}. In particular, we find that standard next-token cross-entropy does not directly provide signal for where and when the model should attend, 
causing models to overlook sign-relevant regions and frames. To address this challenge, we propose \textbf{AttnSign}, a VLM-based spatial-temporal attention steering framework for sign language translation. AttnSign first introduces spatial attention supervision for sign-relevant regions, such as face and hands, in each frame; then develops an RL-based motion-cadence steering method that encourages the model to explore and focus on sign-level keyframes. Experimental results on
How2Sign and OpenASL benchmarks show that our proposed AttnSign consistently outperforms existing methods.
Code will be released.
\end{abstract}

\section{Introduction}\label{sec:introduction}

\begin{figure*}[t]

    \centering
    \includegraphics[width=\textwidth]{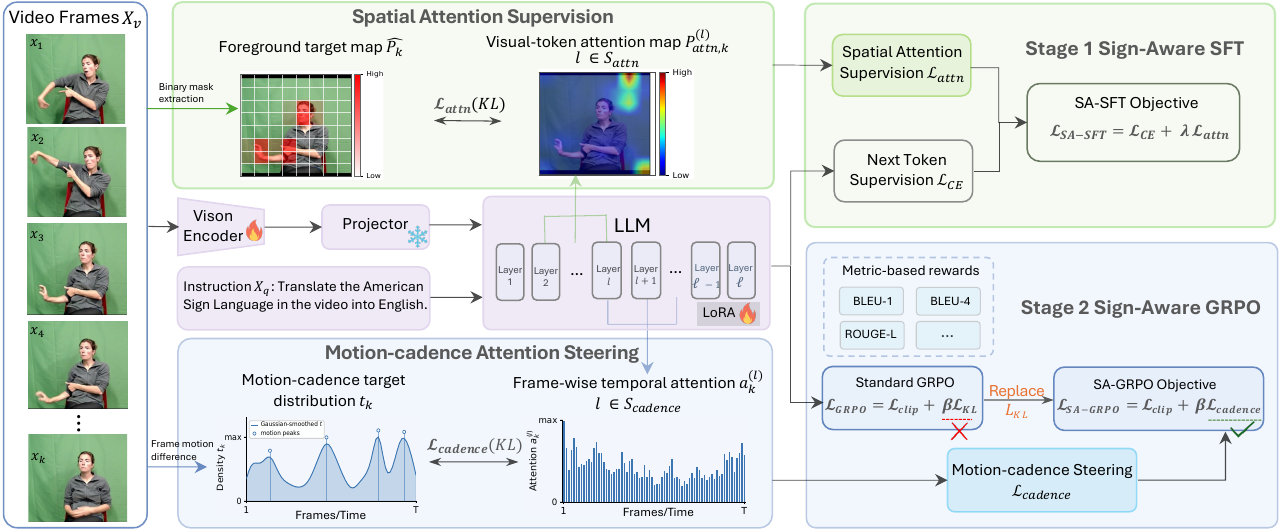}
     \caption{Overview of the proposed training framework. The general VLM takes video frames $X_{v}$ and instruction $X_q$  as inputs and generates the translation. In Sign-Aware SFT stage, we introduce a spatial attention  $\mathcal{L}_{attn}$ at selected LLM layers with frame-level foreground targets, jointly with the next-token prediction loss $\mathcal{L}_{CE}$. In Sign-Aware GRPO stage, we add a  motion-cadence steering objective $\mathcal{L}_{cadence}$ in place of the KL penalty term $\mathcal{L}_{KL}$, which  aligns per-frame temporal attention $a^{(l)}_{k}$ with a Gaussian-smoothed motion-peak target $t_{k}$.  We adopt the translation quality-based rewards.}
    \label{fig:framework}
\end{figure*}

Sign language translation (SLT) converts continuous sign language videos into natural-language sentences. Early systems rely on gloss-based supervision~\cite{camgoz2020sign,Zhang2023SLTUNET}, while more recent methods bypass it entirely~\cite{wong2024sign2gpt,fish2026geo,kim2025leveraging,jiang2026think}. These gloss-free methods typically adopt a modular design: a visual encoder is pretrained with pseudo-gloss supervision~\cite{guo2026bridging}, contrastive video-text alignment~\cite{zhou2023GFSLTVLP}, or self-supervised learning~\cite{hu2023signbert+,gueuwou2025shubert}, and then fine-tuned jointly with a language decoder.

Meanwhile, vision-language models (VLMs) such as Gemini~\cite{team2023gemini}, GPT-4~\cite{achiam2023gpt}, and Qwen-VL~\cite{bai2023qwen} have rapidly advanced, enabling strong general-purpose multimodal perception and generation in a single unified backbone. This naturally raises the question: \emph{can general VLMs serve as effective SLT translators?} However, we empirically observe that these VLMs show limited performance on SLT. We thereby investigate how to close this gap and propose \textbf{AttnSign}, an end-to-end, gloss-free training framework that adapts a general VLM for sign language translation.

To adapt VLMs for SLT, we start with supervised fine-tuning (SFT)~\cite{liu2023visual} on visual instructional data, which teaches the VLM to translate by following instructions. After training, we observe that the model can output fluent translations while assigning high visual attention to background and clothing regions, resulting in poor interpretability.
Yet, sign language conveys linguistic content through manual features such as handshape, movement, and location, as well as non-manual facial markers~\cite{brentari2010handshape,pfau2010nonmanuals}, whereas these non-signing regions carry no linguistic information. In this work, we hypothesize that cross-entropy applied only to text tokens cannot provide explicit signals to regularize misaligned attention. 
We address this problem by adding a spatial attention supervision, enabling the model to generate translations while aligning visual attention with an annotated prior. 

SFT optimizes a token-level objective, while translation quality is judged at the sequence level. To complement SFT, we refine the model as a translation policy via reinforcement learning (RL),  following existing works~\cite{feng2026video}.

Standard RL algorithms such as GRPO~\cite{shao2024deepseekmath} are task-agnostic, treating sign language as any other text-generation task without considering its specific properties. Meanwhile, we observe that its KL penalty term offers limited benefit and can restrict policy updates on our task, consistent with recent RL studies on LLM~\cite{yu2025dapo,hu2025openreasonerzero}. To inject sign-specific signal, we leverage a temporal property of sign language: signs are organized through structured movements and holds that produce changes in visual motion energy over time~\cite{brookshire2017visual}, with motion peaks marking visually salient keyframes within each sign. Given that the KL penalty can offer limited benefit, we then replace it with a motion-cadence steering mechanism that guides temporal attention toward these keyframes during RL training.

We evaluate AttnSign on two large-scale American Sign Language benchmarks, How2Sign and OpenASL. Our proposed AttnSign reaches BLEU-4 of 15.54 on How2Sign and 20.75 on OpenASL, outperforming leading specialized SLT methods and general-purpose VLM baselines. We summarize the contributions as follows:
\begin{itemize}[noitemsep,topsep=2pt]
     \item We propose AttnSign, an end-to-end training framework that adapts a general VLM for SLT, achieving better interpretability and accuracy.
\item In the SFT stage, we introduce a spatial attention supervision that guides the model to focus on sign-relevant regions in each frame.
\item In the RL stage, we propose a motion-cadence steering mechanism in place of the KL penalty in standard GRPO, guiding the model's temporal attention toward keyframes of each sign.
    \item Experiments on benchmark datasets show the leading performance over both task-specific methods and general-purpose VLMs. 
   
\end{itemize}

\section{Related Work}\label{sec:related}

\textbf{Sign Language Translation.}
Prior SLT methods use gloss annotations to bridge visual and linguistic modalities~\cite{Chen2022MMTLB,Zhang2023SLTUNET}, which bring promising performance but can be hard to scale due to the annotation cost. More recently, gloss-free methods~\cite{zhou2023GFSLTVLP,gong2024llms,rust2024towards,pu2026signdpo,hamidullah2024sign,karakucs2026semantic} learn from video-text pairs  with auxiliary alignment or pretraining objectives, being more flexible and seeing further performance boost. Given the success of foundation models~\cite{liu2020multilingual,touvron2023llama, bai2023qwen}, existing works~\cite{zhang2024scaling,wong2024sign2gpt,hwang2024efficient,rao2025rvlf,jang2025lost}  adopt LLMs for decoding. However, these works generally need to tailor the model structure by integrating customized vision encoders~\cite{he2016deep,oquab2023dinov2}. How to directly tune a unified VLM model for SLT remains underexplored.

\textbf{Training Strategies for SLT.}
These SLT systems are typically trained in two stages: visual-encoder pretraining followed by joint fine-tuning. Pretraining objectives include contrastive alignment between video and text~\cite{kim2025leveraging}, masked self-supervised modeling~\cite{rust2024towards,gueuwou2025shubert}, and pseudo-gloss prediction~\cite{wong2024sign2gpt}. Most works rely on SFT for the fine-tuning stage~\cite{liang2024llava,gong2024llms,zhang2024scaling}, while recent works~\cite{rao2025rvlf,pu2026signdpo} further explore reinforcement learning after SFT. Meanwhile, general-purpose VLMs—including LLaVA-OneVision~\cite{li2024llava} and Qwen-VL~\cite{bai2025qwen2} offer strong general-purpose visual-language capabilities~\cite{zhao2026cnsl,inan2025signalignlm}. Instead of relying on a separate pretraining stage, we adapt a general VLM end-to-end through SFT and RL with sign-specific domain knowledge introduced at both stages.

\textbf{Attention Supervision.}
Attention supervision has been studied in vision and multimodal learning to improve grounding, including attention transfer~\cite{li2024surprising}, human attention or gaze guidance~\cite{selvaraju2019hint,yan2024voila}, and region-level grounding for vision-language models~\cite{wan2024contrastive}. These works demonstrate that attention supervision can effectively shape model behavior, yet its potential remains underexplored in sign language translation. To address this gap, we develop attention steering methods tailored to SLT by guiding attention in the LLM decoder along both spatial and temporal dimensions.

\section{Method}\label{sec:method}

Section~\ref{subsec:vlm} introduces SFT on a general VLM, and Section~\ref{subsec:attn_sft} extends it with spatial attention supervision. Section~\ref{subsec:grpo} describes GRPO for sequence-level optimization, and Section~\ref{subsec:sa_grpo} enhances it with motion-cadence attention steering.

\subsection{Instructional SFT on a VLM}\label{subsec:vlm}
Given a sign language video $X_v = \{x_k\}_{k=1}^K$ of $K$ frames, language instruction $X_q$, and a target translation $Y$, the goal is to learn a translator $\pi_\theta$ parameterized by $\theta$, such that $Y = \pi_\theta(X_v,X_q)$. 

We use a general VLM that models perception and generation of each output token $y_n\in\{y_1\cdots y_N\}=Y$ jointly as:
\begin{equation}
    P_\theta(Y \mid X_v, X_q) = 
    \prod_{n=1}^{N} P_\theta\!\left(y_n \mid X_v, X_q, y_{<n}\right)\,.
\end{equation}
where the input frames $X_v$ are encoded into visual tokens using a vision encoder and projected to language space by a projector, followed by the instruction tokens.  

\textbf{Instructional Fine-tuning.}
Pretrained VLMs provide three capabilities directly useful for SLT: general visual perception, fluent language generation, and alignment between visual tokens and language. However, they can lack SLT-specific knowledge due to the scale and quality of the sign language data in the real world~\cite{duarte2021how2sign}. We empirically find that representative pretrained VLMs such as GPT-5 and Qwen2.5-VL-7B~\cite{bai2025qwen2} can show limited performance on both How2Sign and OpenASL (Table~\ref{tab:main_h2s} and Table~\ref{tab:openasl}).

We cast SLT as instruction-following generation task. Each training instance pairs a video $X_v$ with a natural-language instruction $X_q$ (\emph{e.g.}, ``\textit{Translate the American Sign Language in this video to English}'') and uses the target translation $Y$ as the response. The model is trained with the standard cross-entropy loss:
\begin{equation}
    \mathcal{L}_{\mathrm{CE}} = -\sum_{n=1}^{N} \log P_\theta(y_n \mid X_v, X_q,y_{<n})\,,
\end{equation}
where $\theta$ is optimized to maximize the likelihood of the translation conditioned on the input video and instruction. The SFT stage transfers the VLM's general capabilities to the SLT task.

\begin{figure}[t]
    \centering
    \includegraphics[width=\columnwidth]{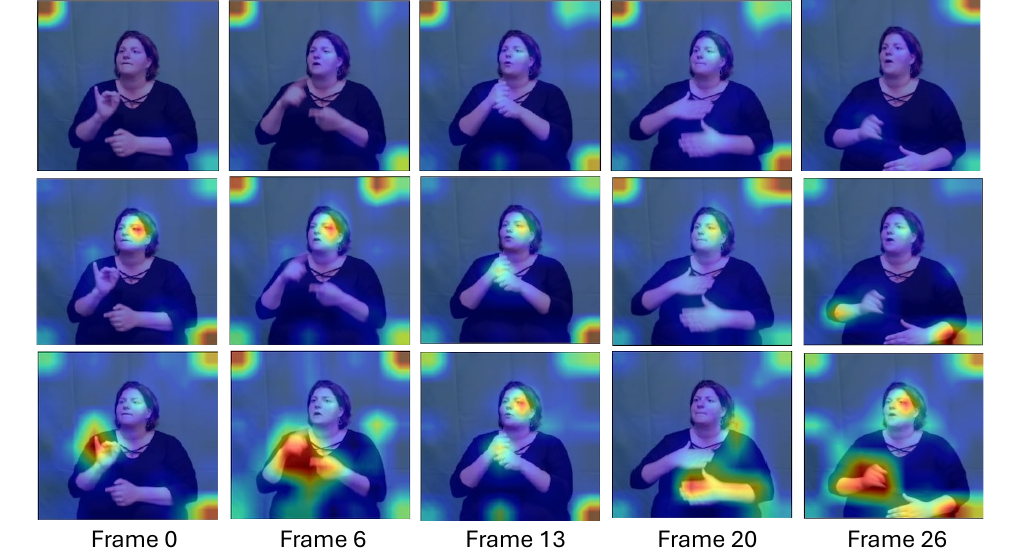}
     \caption{Visual-token attention maps at LLM layer 16 across five frames (columns) of the same test video. Rows show the three training stages: pre-trained VLM (top), standard SFT (middle), and Sign-Aware SFT with our spatial attention supervision (bottom).}
    \label{fig:attn_heatmap_analysis}
    \vspace{-3mm}
\end{figure}

\textbf{Observation on Visual Attention.}
Despite the fact that sign-linguistic content is primarily conveyed by the signer's hands and face, We empirically find that after instructional SFT, the model can assign high visual attention to the less informative areas, including background and clothing regions (Figure~\ref{fig:attn_heatmap_analysis}, middle column). Considering that cross-entropy on text tokens does not penalize misaligned attention, we hypothesis that applying explicit spatial constraint to attend more on sign-relevant regions can benefit translation.

\subsection{SA-SFT: Sign-Aware SFT}\label{subsec:attn_sft}
To achieve the goal, we propose a spatial supervision loss during SFT. Unlike approaches that rely on cropped regions or keypoints as inputs~\cite{gan2023realtime,gueuwou2025signmusketeers,li2025uni}, we keep the raw video as input and impose the spatial constraint through the loss. This preserves global visual context while guiding attention toward hand and face regions that carry primary sign-linguistic cues.

\textbf{Foreground target.}
For each frame $x_k$ of a training video, we use MediaPipe~\cite{lugaresi2019mediapipe} to extract a binary mask $\mathbf{M}_k$ covering the hands and face. We then resize the mask via average pooling to the $8{\times}8$ visual-token grid, yielding a soft foreground intensity vector $\mathbf{M}_{k,\mathrm{flat}} \in [0,1]^{64}$, and convert it into a soft spatial target using a temperature-controlled softmax:
\begin{equation}
    \hat{P}_k = \mathrm{softmax}(\mathbf{M}_{k,\mathrm{flat}} / T).
\end{equation}

\textbf{Spatial Attention loss.}
Let $P^{(l)}_{\mathrm{attn},k} \in \mathbb{R}^{64}$ denote the per-frame attention distribution over the 64 visual positions of frame $k$ at LLM decoder layer $l$. We obtain this distribution by extracting the attention weights assigned to visual tokens and renormalizing them within each frame, so that the 64 spatial positions of frame $k$ form a valid probability distribution. We supervise a set of decoder layers $\mathcal{S}_{\mathrm{attn}}$ with:
\begin{equation}
    \mathcal{L}_{\mathrm{attn}} =
    \frac{1}{K\,|\mathcal{S}_{\mathrm{attn}}|}
    \sum_{k=1}^{K}
    \sum_{l \in \mathcal{S}_{\mathrm{attn}}}
    \mathrm{KL}\!\left(P^{(l)}_{\mathrm{attn},k} \;\|\; \hat{P}_k\right),
\end{equation}
averaging over frames and supervised layers.

This objective penalizes attention mass assigned to regions with low target probability, thereby encouraging the model to concentrate attention on the hand and face regions. The SA-SFT objective combines the standard next-token cross-entropy loss with spatial attention supervision:
\begin{equation}
    \mathcal{L}_{\mathrm{SA\text{-}SFT}} = \mathcal{L}_{\mathrm{CE}} + \gamma\,\mathcal{L}_{\mathrm{attn}},
\end{equation}
where $\gamma$ controls the strength of attention supervision.
We supervise mid-depth LLM decoder layers, which we empirically find to be the most effective for spatial attention supervision (Section~\ref{subsec:discussion}).

SFT optimizes token-level cross-entropy under teacher forcing, whereas translation quality is evaluated with sequence-level metrics such as BLEU, ROUGE, and BLEURT. This creates an objective mismatch: cross-entropy encourages high-probability next-token predictions, but does not directly optimize full-sentence quality or enhance the semantic score. Following existing works~\cite{feng2026video}, we therefore further perform  post-training   stage upon SFT model, \emph{e.g.}, policy optimization. 

\subsection{GRPO for Sequence-Level Optimization}\label{subsec:grpo}
 We first study the canonical GRPO~\cite{shao2024deepseekmath} to further fine-tune the SFT model for the SLT, considering GRPO can also support different rule-based rewards for optimizing non-differentiable translation metrics.

\textbf{Formulation.}
For each input video $X_v$, we sample $G$ candidate translations $\{Y^{(g)}\}_{g=1}^G$ from the current policy and score each with a reward $r(X_v, Y^{(g)})$. 
We define the reward as a reference-based function:
\begin{equation}
    r(X_v,Y^{(g)}) = R(Y^{(g)}, Y^\ast),
\end{equation}
where $Y^\ast$ is the reference translation and $R(\cdot,\cdot)$ measures translation quality. This formulation allows us to instantiate $R$ with different rule-based reward designs, which we compare in Section~\ref{sec:experiments}.

We compute group-relative advantages by comparing each candidate with the group mean reward:
\begin{equation}
    A^{(g)} = r(X_v, Y^{(g)}) - \frac{1}{G}\sum_{g'=1}^{G} r(X_v,Y^{(g')}).
\end{equation}
The advantage $A^{(g)}$ is assigned to all generated tokens of $Y^{(g)}$.
The standard GRPO objective is defined as:
\begin{equation}
    \mathcal{L}_{\mathrm{GRPO}}
    =
    \mathcal{L}_{\mathrm{clip}}
    +
    \beta\,\mathcal{L}_{\mathrm{KL}},
\end{equation}
where $\mathcal{L}_{\mathrm{clip}}$ is the clipped GRPO surrogate, $\beta$ controls the KL strength, and $\mathcal{L}_{\mathrm{KL}}$ keeps the policy close to a fixed reference $\pi_{\mathrm{ref}}$. Let $c=(X_v,X_q)$. The KL penalty term is:
\begin{equation}
    \mathcal{L}_{\mathrm{KL}}
    =
    \mathbb{E}_{c\in D}
    \left[
    D_{\mathrm{KL}}\!\left(
        \pi_\theta(\cdot \mid c)
        \middle\|
        \pi_{\mathrm{ref}}(\cdot \mid c)
    \right)
    \right],
\end{equation}
where $D_{\mathrm{KL}}$ is the KL divergence.

\textbf{Limitations of standard GRPO.}
Standard GRPO can have limitations for SLT. First, it is task-agnostic and carries no explicit SLT-specific knowledge. Second, we observe that its generic $\mathcal{L}_{\mathrm{KL}}$, which only anchors the policy to the reference $\pi_{\mathrm{ref}}$, can bring limited benefit and even can restrict the potential of the policy model for SLT (Table~\ref{tab:component_ablation}). This is consistent with recent RL studies on LLM~\cite{yu2025dapo,hu2025openreasonerzero}. Accordingly we replace $\mathcal{L}_{\mathrm{KL}}$ with a proposed sign-specific steering mechanism that addresses both issues.

\subsection{SA-GRPO: Sign-Aware GRPO}\label{subsec:sa_grpo}
 Sign language exhibits structured temporal dynamics, with signs organized through movements and holds that give rise to changes in visual motion energy over time~\cite{liddell1989american,brookshire2017visual}. This makes motion peaks a natural target for guiding optimization. However, {this prior alone does not tell us whether focusing attention on these peaks actually helps translation quality, or in which LLM layers this effect holds}. Accordingly, we uncover the attention-performance relationship through a  statistical analysis, connecting the model's per-layer attention concentration with translation quality.

\begin{figure}[t]
    \centering
    \includegraphics[width=\columnwidth]{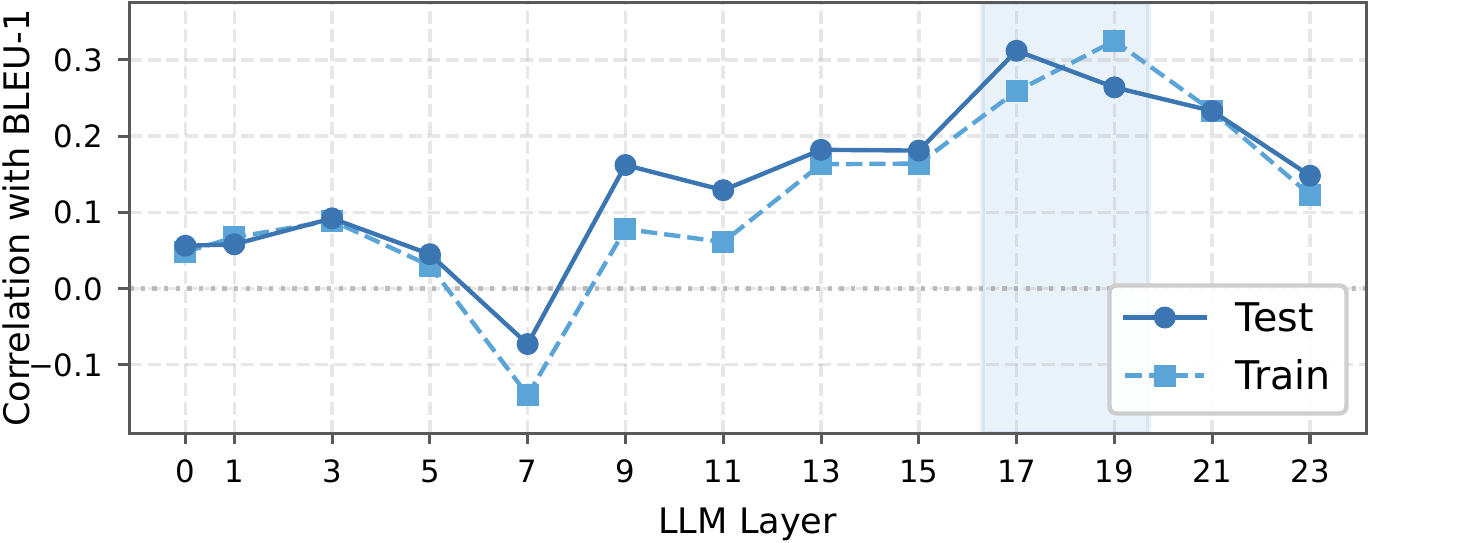}
    \caption{
    Correlation between \textbf{\textit{temporal attention score}} and \textbf{\textit{translation performance}} (measured by BLEU-1). For example, the strongest positive correlation emerges at the 17th and 19th layers for the SA-SFT model, motivating us to steer the temporal attention for the performance boost during optimization. 
    }
    \label{fig:gini_corr}
    \vspace{-3mm}
\end{figure}

\textbf{Statistical Analysis.}
 For each video and LLM layers of the SA-SFT model, we obtain a per-frame attention vector by aggregating over visual tokens within each frame and averaging over text-query positions. We compute the concentration score as $C = 1 - H/\log K$, where $H$ is the entropy over $K$ frames. Figure~\ref{fig:gini_corr} shows that there exists a positive correlation relationship between $C$ and BLEU-1, computed across both training and test videos, especially peaks at the upper-middle 17th and 19th layers ($r \approx +0.32$), indicating that enhancing such an attention concentration at these layers \textit{during optimization} can potentially  improve the translation quality \textit{during inference}.

\textbf{Motion-cadence Steering Loss.}
While the analysis identifies at which layers concentration helps, the linguistic prior that informative moments cluster around motion peaks tells us which frames to concentrate on. We therefore construct a frame-level target around motion peaks at these layers. For each training video, we compute a per-frame motion signal using grayscale frame differences:
\begin{equation}
    m_k = \frac{1}{HW}\sum_{h,w}
    \left|x_k^{\mathrm{gray}} - x_{k-1}^{\mathrm{gray}}\right|_{h,w},
\end{equation}
where $x_k^{\mathrm{gray}}$ denotes the grayscale frame at time $k$. We detect local maxima $\mathcal{P}$ in $m_k$ with a minimum inter-peak distance of 3 frames, and convert them into a normalized temporal target:
\begin{equation}
    t_k = \frac{1}{Z}\sum_{p \in \mathcal{P}}
    \exp\!\left(-\frac{(k-p)^2}{2\sigma^2}\right),
    \qquad \sum_k t_k = 1.
\end{equation}
The Gaussian smoothing spreads each peak to neighboring frames, yielding a soft cadence target.

% \textbf{Motion-cadence attention steering.}
Let $a^{(l)}_k$ denote the attention mass on frame $k$ at LLM layer $l$. We obtain it by summing attention over the 64 visual tokens of frame $k$ and normalizing across all frames. We supervise a set of layers $\mathcal{S}_{\mathrm{cadence}}$ toward this target with:
\begin{equation}
    \mathcal{L}_{\mathrm{cadence}} =
    \frac{1}{|\mathcal{S}_{\mathrm{cadence}}|}
    \sum_{l \in \mathcal{S}_{\mathrm{cadence}}}
    \mathrm{KL}\!\left(a^{(l)} \;\|\; t\right).
\end{equation}

This loss keeps temporal attention aligned with the soft cadence target $t$ centered on motion peak.

\textbf{SA-GRPO.}
We replace the $\mathcal{L}_{\mathrm{KL}}$ in standard GRPO with the sign-specific cadence term $\mathcal{L}_{\mathrm{cadence}}$. The SA-GRPO objective is:
\begin{equation}
    \mathcal{L}_{\mathrm{SA\text{-}GRPO}}
    =
    \mathcal{L}_{\mathrm{clip}}
    +
    \lambda\,\mathcal{L}_{\mathrm{cadence}},
\end{equation}
where $\lambda$ controls the strength of the cadence steering. Unlike generic $\mathcal{L}_{\mathrm{KL}}$, $\mathcal{L}_{\mathrm{cadence}}$ is conditioned on the input video and ties the regularization to sign-relevant motion structure.
Interestingly, we observe a continued performance boost tendency by reducing the KL penalty in canonical GRPO or increasing our proposed motion-cadence term. See more discussion in  Section~\ref{subsec:discussion}.

\begin{table*}[t]
\centering
\footnotesize
\renewcommand{\arraystretch}{1.15}
\definecolor{ourscolor}{HTML}{DEEBF7} 
\definecolor{typecolor}{HTML}{0000FF}
\setlength{\tabcolsep}{11pt}

\begin{tabular}{lcccccc}
\hline
Method & B-1 & B-2 & B-3 & B-4 & ROUGE-L & BLEURT \\
\hline

\rowcolor{white}
\textcolor{typecolor}{\textit{Closed-source VLMs}} & & & & & & \\
Gemini-2.5-Pro~\cite{comanici2025gemini} & 15.41 & 5.57 & 2.61 & 1.41 & 13.79 & 30.90 \\

GPT-5                                  & 6.64  & 1.02 & 0.68 & 0.10 & 5.72  & 24.62 \\
\hline 

\rowcolor{white}
\textcolor{typecolor}{\textit{Open-source VLMs}} & & & & & & \\
LLaVA-OV-7B~\cite{li2024llava}           & 11.21 & 3.56 & 1.45 & 0.16 & 4.10 & 20.15 \\
Qwen2.5-VL-7B~\cite{bai2025qwen2}        & 12.42 & 4.12 & 1.13 & 0.11 & 7.24 & 24.10 \\
InternVL-2.5-8B~\cite{chen2024expanding} & 11.81 & 2.35 & 0.86 & 0.06 & 7.56 & 25.12 \\
\hline

\rowcolor{white}
\textcolor{typecolor}{\textit{Trained on H2S only}} & & & & & & \\
GloFE-VN~\cite{lin2023gloss}                  & 14.94 & 7.27 & 3.93 & 2.24 & 12.61 & 31.65 \\
Tarr\'es et al.~\cite{tarres2023sign}         & 34.01 & 19.30 & 12.18 & 8.03 & ---   & ---   \\
SSVP-SLT~\cite{rust2024towards} & 30.20 & 16.70 & 10.50 & 7.00 & 25.70 & 39.30 \\

SignMusketeers~\cite{gueuwou2025signmusketeers} & 18.80 & 8.10 & 4.20 & 2.40 & --- & --- \\
C$^2$RL~\cite{chen2025c}                      & 29.07 & 18.56 & 12.92 & 9.37 & 27.02 & ---   \\
\rowcolor{gray!15}
\textbf{AttnSign (Ours)}                      & \textbf{35.82} & \textbf{20.68} & \textbf{13.52} & \textbf{9.45} & \textbf{30.45} & \textbf{42.73} \\
\hline

\rowcolor{white}
\textcolor{typecolor}{\textit{Trained on H2S + YT}} & & & & & & \\
Uthus et al.~\cite{uthus2023youtube}   & 37.82 & 24.13 & 16.92 & 12.39 &  ---  & 46.63   \\
SSVP-SLT-LSP~\cite{rust2024towards}    & 43.20 & 28.8 & 20.80 & 15.50 & 38.40 & 49.60 \\
SignMusketeers~\cite{gueuwou2025signmusketeers} & 41.50 & 27.20 & 19.30 & 14.30 & --- & --- \\
Uni-Sign~\cite{li2025uni}              & 40.20 & --- & --- & 14.90 & 36.00 & 49.40 \\
\rowcolor{gray!15}
\textbf{AttnSign (Ours)}               & \textbf{43.45} & \textbf{28.82} & \textbf{20.94} & \textbf{15.54} & \textbf{39.02} & \textbf{50.44} \\
\hline

\end{tabular}

\caption{Performance comparison on the How2Sign benchmark. Methods are grouped into four blocks: zero-shot closed-source VLMs, zero-shot open-source VLMs, specialized SLT methods using only How2Sign (H2S) dataset, and specialized SLT methods that additionally use YouTube-ASL (YT) dataset. B-$n$ denotes BLEU-$n$. Our method uses InternVL-2.5-1B as the backbone.}
\label{tab:main_h2s}
\end{table*}
\begin{table}[t]
\centering
\footnotesize
\renewcommand{\arraystretch}{1.15} 
\definecolor{ourscolor}{HTML}{DEEBF7}
\definecolor{typecolor}{HTML}{0000FF} 
\setlength{\tabcolsep}{4pt}

\resizebox{\columnwidth}{!}{%
\begin{tabular}{lcccc}
\hline
Method & B-1 & B-4 & ROUGE-L & BLEURT \\
\hline

\rowcolor{white}
\textcolor{typecolor}{\textit{Closed-source VLMs}} & & & & \\
Gemini-2.5-Pro~\cite{comanici2025gemini} & 22.23 & 3.82 & 19.78 & 36.45 \\
GPT-5    & 14.18 & 1.63 & 10.23 & 28.36 \\
\hline

\rowcolor{white}
\textcolor{typecolor}{\textit{Open-source VLMs}} & & & & \\
Qwen2.5-VL-7B~\cite{bai2025qwen2}   & 11.38 & 0.55 & 6.30 & 24.71 \\
InternVL-2.5-8B~\cite{chen2024expanding} & 12.96 & 0.38 & 6.09 & 21.15 \\
\hline

\rowcolor{white}
\textcolor{typecolor}{\textit{Specialized SLT methods}} & & & & \\

I3D-transformer~\cite{shi2022open}  & 18.31 & 5.66  & 18.64 & 28.82 \\
OpenASL~\cite{shi2022open}   & 20.92 & 6.72  & 21.02 & 31.09 \\
GloFE-VN~\cite{lin2023gloss} & 21.56 & 7.06  & 21.75 & 36.35 \\
C$^{2}$RL~\cite{chen2025c} & 31.46 & 13.21 & 31.36 & ---   \\

\rowcolor{gray!15}
\textbf{AttnSign (Ours)}               & \textbf{37.52}   & \textbf{13.45}   & \textbf{34.23}   & \textbf{51.42}   \\
\rowcolor{gray!15}
\textbf{AttnSign (Ours)}$^{\dagger}$    & \textbf{47.74} & \textbf{20.75} & \textbf{45.62} & \textbf{61.68} \\
\hline

\end{tabular}
}

\caption{Performance comparison on the OpenASL benchmark. Methods are grouped into three blocks: zero-shot closed-source VLMs, zero-shot open-source VLMs, and specialized SLT methods. $^{\dagger}$: additionally uses YouTube-ASL data. B-$n$ denotes BLEU-$n$. Our method uses InternVL-2.5-1B as the backbone.}
\label{tab:openasl}
\end{table}

\section{Experiments}\label{sec:experiments}
\subsection{Datasets and Metrics}\label{subsec:datasets}

\textbf{Datasets.} We train and evaluate on two large-scale American Sign Language (ASL) benchmarks: How2Sign (H2S)~\cite{duarte2021how2sign}, about 80 hours of studio-recorded ASL, and OpenASL~\cite{shi2022open}, about 288 hours ASL videos. For SFT scaling, we additionally incorporate YouTube-ASL (YT)~\cite{uthus2023youtube}, a 984-hour web-scale corpus with weakly aligned English text translations. For GRPO training, we use only the in-domain training split of the target benchmark.

\textbf{Evaluation metrics.} Following previous work      
~\cite{rust2024towards,lin2023gloss} we report BLEU ~\cite{papineni2002bleu} computed with SacreBLEU~\cite{post2018call}, ROUGE-L~\cite{lin2004rouge}, and BLEURT~\cite{sellam2020bleurt}.

\subsection{Implementation Details}\label{subsec:impl}
 Our backbone is InternVL-2.5-1B~\cite{chen2024internvl}. Each video frame is resized to $224{\times}224$ and frames are uniformly sampled over time at 16~FPS. Throughout training and inference, the model takes raw video and instructions as input.

\textbf{SFT stage.} We fully fine-tune the ViT and adapt the LLM with LoRA~\cite{hu2021lora} at rank $r{=}16$. We use AdamW with a learning rate of $5{\times}10^{-5}$ and cosine decay, training for up to 5 epochs with a batch size of 64 on $8\times$ NVIDIA A100 GPUs, taking approximately 12 hours in the H2S-only setting and 6 days in the full H2S+YT setting. For spatial attention supervision, we use weight $\gamma{=}0.15$, softmax temperature $T{=}0.5$, supervise layers 13 and 16 of LLM. Pose-based foreground masks are extracted only for How2Sign and OpenASL samples; YouTube-ASL samples are included in SFT with the cross-entropy loss only, and attention supervision is skipped on these samples.

\textbf{RL stage.} We start from the SA-SFT model, with group size $G{=}4$, clipping $\epsilon{=}0.2$, and sampling temperature 1.0. As in the SFT stage, we fully fine-tune the ViT and adapt the LLM with LoRA at rank $r{=}16$. Training runs for 2 epochs on the 10K in-domain samples take approximately 8 hours. For How2Sign evaluation, we use the How2Sign training split, and for OpenASL evaluation, we use the OpenASL training split. We use BLEU-1 and ROUGE-L as the rule-based rewards with equal weight. The motion-cadence attention constraint uses $\lambda{=}0.001$ and is applied at LLM layers 17 and 19.

\subsection{Main results}\label{subsec:quantitative}

Tables~\ref{tab:main_h2s} and \ref{tab:openasl} report our main results on How2Sign and OpenASL. On both benchmarks and under both the single-source and +YT augmentation training settings, AttnSign achieves the best performance on every metric, outperforming the leading specialized baselines and general-purpose VLMs. Among the baselines, SSVP-SLT~\cite{rust2024towards} uses the same H2S-only and H2S+YT training settings as ours but also includes a self-supervised pretraining stage on YT dataset. AttnSign exceeds it under both settings without that additional pretraining stage, reaching BLEU-4 of $9.45$ on H2S-only compared to $7.00$, and $15.54$ on H2S+YT compared to $15.50$. AttnSign also leads on ROUGE-L and BLEURT under both settings. Together, these results indicate that a general VLM with domain-specific supervision at each training stage achieves competitive performance with prior methods.

\subsection{Ablation Study}\label{subsec:ablation}

We validate the contribution of each pipeline component. Throughout this section, SFT is trained on How2Sign and YouTube-ASL, GRPO is trained on 10K How2Sign in-domain samples, and evaluation is on the How2Sign. Appendix~\ref{app:openasl_ablation} reports the same ablation on OpenASL with matching gains.

\textbf{Effect of each component.}
Table~\ref{tab:component_ablation} reports the contribution of each pipeline component. Instructional SFT raises performance over the zero-shot baseline, and spatial attention supervision in SA-SFT further improves BLEU-4 from $13.89$ to $14.55$ over plain SFT. Within the RL stage, removing the KL term of GRPO improves BLEU-4 from $15.04$ to $15.38$, indicating that anchoring the policy to the SFT reference limits beneficial updates. Combining motion-cadence steering with the KL penalty (w/ KL + cadence) reaches $15.13$. SA-GRPO, which replaces the KL penalty with motion-cadence steering, reaches $15.54$. This further outperforms GRPO w/o KL by $+0.16$ BLEU-4, indicating that cadence steering provides task-aligned guidance beyond what removing KL alone offers. This supports replacing the KL penalty with cadence steering rather than combining the two.

\begin{table}[t]
\centering
\small
\setlength{\tabcolsep}{4pt}
\resizebox{\columnwidth}{!}{%
\begin{tabular}{lcccc}
\toprule
Method & B-1 & B-4 & ROUGE-L & BLEURT \\
\midrule
Zero-shot baseline          & 10.79    & 0.08    & 6.98    & 23.87    \\
\midrule
\multicolumn{5}{l}{\textit{SFT stage}} \\
\quad SFT                            & 41.23   & 13.89 & 37.13 & 49.02 \\

\rowcolor{gray!15}
\quad SA-SFT (Ours)                     & 41.94    & 14.55 & 37.67 & 49.55 \\

\midrule
\multicolumn{5}{l}{\textit{RL stage}} \\
\quad GRPO (w/ KL)        & 42.74    & 15.04 & 38.47 & 50.01 \\
\quad GRPO (w/o KL)       & 43.22    & 15.38 & 38.86 & 50.30 \\
\quad GRPO (w/ KL + cadence)       & 42.87    & 15.13 & 38.62 & 50.13 \\
\rowcolor{gray!15}
\quad SA-GRPO (Ours)      & \textbf{43.45} & \textbf{15.54} & \textbf{39.02} & \textbf{50.44} \\
\bottomrule
\end{tabular}
}

\caption{Component ablation on How2Sign. We isolate the contribution of each stage in our pipeline: the SFT stage block compares standard SFT with our SA-SFT, and the RL stage block compares variants of GRPO. Combining SA-SFT and SA-GRPO yields the best results across all metrics.}
\label{tab:component_ablation}
\end{table}

\subsection{Model Discussion}\label{subsec:discussion}
All analyses in this section are evaluated on How2Sign. Unless otherwise stated, the training data is also How2Sign only.

\begin{figure}[t]
\centering
\includegraphics[width=\columnwidth]{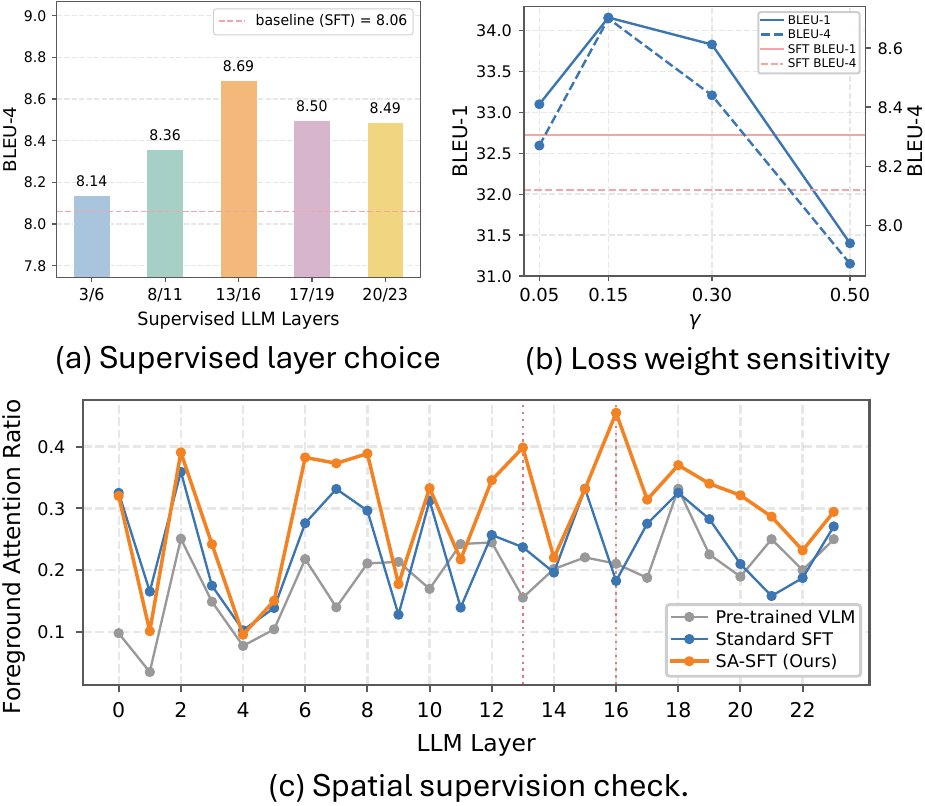}
\caption{SA-SFT analysis. (a) Layer choice: sweep over supervised layer pairs spanning the LLM depth; mid-depth performs best in BLEU-4. (b) Spatial attention loss weight $\gamma$ sensitivity. (c) Per-layer foreground attention ratio: SA-SFT raises mass on foreground tokens at the supervised 13th and 16th layers over standard SFT and the pre-trained VLM.}
\label{fig:sft_attn_ablations}
\vspace{-2mm}
\end{figure}

\textbf{Layer choice for spatial supervision.}
We vary the supervised layer set $\mathcal{S}_{\mathrm{attn}}$ while fixing $\gamma$ at its default. Figure~\ref{fig:sft_attn_ablations}(a) sweeps supervised layer pairs spanning the LLM depth. Mid-depth supervision at the 13th and 16th layers produces the strongest gains, raising BLEU-4 from the baseline $8.06$ to $8.69$, while shallow, upper-middle, and final supervision yield smaller or no improvement. Mid-depth supervision consistently outperforms shallow or final layers across all metrics, supporting our choice of the 13th and 16th layers as default.

\textbf{Spatial supervision verification and sensitivity.}
Fixing the supervised layers at the 13th and 16th, Figure~\ref{fig:sft_attn_ablations}(b) sweeps the loss weight $\gamma$: the curve shows a clear interior optimum at $\gamma{=}0.15$, with smaller values weakening supervision and larger values over-regularizing. 

Figure~\ref{fig:sft_attn_ablations}(c) reports the foreground attention ratio—the fraction of per-frame attention mass falling inside the pose-derived foreground mask—across LLM layers. SA-SFT raises this ratio at the supervised 13th and 16th layers compared with standard SFT and the pre-trained VLM, confirming that the supervision actually shifts attention toward the foreground regions.

\textbf{GRPO reward design.}
\begin{table}[t]
\centering
\small
\setlength{\tabcolsep}{4pt}
\resizebox{\columnwidth}{!}{%
\begin{tabular}{lcccc}
\toprule
Reward & B-1 & B-4 & ROUGE-L & BLEURT \\
\midrule
SA-SFT (baseline) & 34.16 & 8.69 & 29.31 & 42.21 \\
\midrule
BLEU-1   & 35.03 & 9.18 & 30.21 & 42.42 \\
BLEU-1 + BLEU-4 & 34.78 & 8.98 & 29.89 & 42.33 \\
\rowcolor{gray!15}
BLEU-1 + ROUGE-L & \textbf{35.68} & \textbf{9.28} & \textbf{30.33} & \textbf{42.62} \\
BLEU-1 + BLEU-4 + ROUGE-L  & 35.47 & 9.22 & 30.25 & 42.53 \\
\hline
\end{tabular}
}
\caption{Rule-based reward combinations for standard GRPO. The selected reward is reused in SA-GRPO for our main experiments.}
\label{tab:grpo_reward}
\end{table}
Table~\ref{tab:grpo_reward} compares four rule-based reward combinations under standard GRPO, isolating reward design from the motion-cadence steering. BLEU-1 alone improves over the SA-SFT baseline. Adding BLEU-4 consistently degrades performance, suggesting that the sparse 4-gram signal is noisy at this scale. BLEU-1+ROUGE-L performs best across all metrics, and is adopted as the default reward in both standard GRPO and SA-GRPO for our main experiments.

\textbf{SA-GRPO analysis.}
\begin{figure}[t]
\centering
\includegraphics[width=\columnwidth]{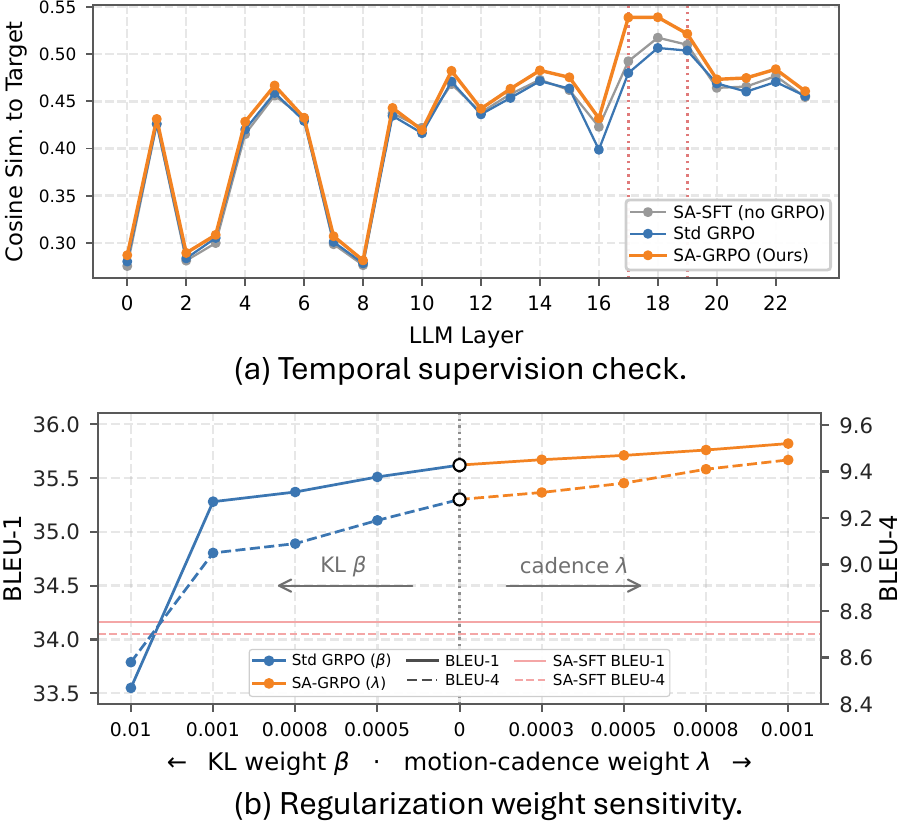}
\caption{(a) Per-layer cosine similarity between attention and the motion-cadence target: SA-GRPO raises alignment at the supervised layers over the SA-SFT model and the standard GRPO model. (b) Sensitivity to the regularization weight; the shared center is the no-regularization baseline. Left: GRPO with KL weight $\beta$. Right: SA-GRPO with motion-cadence weight $\lambda$.}
\label{fig:grpo_reg}
\vspace{-3mm}
\end{figure}
Figure~\ref{fig:grpo_reg}(a) verifies SA-GRPO's mechanism: cosine alignment between attention and the motion-cadence target rises at the supervised 17th and 19th layers, which are the layers identified by the statistical analysis, while GRPO w/o KL remains close to the SA-SFT model. This indicates that the alignment is a specific effect of motion-cadence steering, not a generic outcome of RL training. 

Figure~\ref{fig:grpo_reg}(b) contrasts the two forms of attention control from a shared center--GRPO without KL penalty or cadence term. For standard GRPO, any nonzero KL weight $\beta$ reduces BLEU-4, and the largest weight $\beta{=}0.01$ drops the BLEU-4 by nearly a full point below the center baseline; this confirms that anchoring the policy to $\pi_{\mathrm{ref}}$ limits how much the policy can improve under our setting. Replacing the KL penalty with our motion-cadence steering reverses the pattern: BLEU-4 increases monotonically with $\lambda$ across the swept range, peaking at $\lambda{=}0.001$, indicating that sign-specific steering carries useful signal.

\begin{figure}[t]
\centering
\includegraphics[width=0.96\columnwidth]{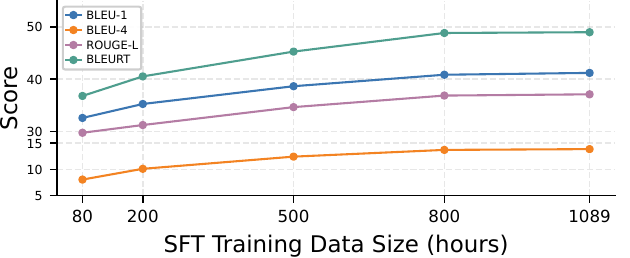}
\caption{Effect of SFT training data size on the H2S, measured by total video duration. The 80h setting uses H2S only; larger sizes are obtained by adding YT data.}
\label{fig:abl_scaling}
\end{figure}

\textbf{SFT Data Scaling}
Figure~\ref{fig:abl_scaling} shows that increasing the SFT training data, starting from the 80h H2S-only setting and progressively adding YT data, yields substantial gains in the early scale-up regime, while further scaling at the larger end yields only marginal additional improvement.
\section{Conclusion}\label{sec:conclusion}

We presented AttnSign, an end-to-end training framework that adapts a general-purpose VLM to SLT in two stages. In the SFT stage, we introduce a spatial attention supervision that guides the model to focus on sign-relevant regions in each frame; in the RL stage, we add a motion-cadence steering mechanism in place of the KL penalty in standard GRPO, guiding the model's temporal attention toward keyframes of each sign. AttnSign achieves competitive performance on both How2Sign and OpenASL, showing that explicitly steering visual attention toward \emph{where} and \emph{when} sign-linguistic content occurs is a useful strategy for adapting a general-purpose VLM to SLT. We hope these findings encourage exploring attention-level supervision for other video-language tasks.

\paragraph{Limitations.} Our experiments focus on American Sign Language (ASL); extending the proposed framework to other sign languages is left for future work. In addition, while attention steering yields strong empirical gains, a theoretical derivation of why this form of supervision benefits translation quality remains an open question that warrants further study.

\bibliography{main}

@article{achiam2023gpt,
  title={Gpt-4 technical report},
  author={Achiam, Josh and Adler, Steven and Agarwal, Sandhini and Ahmad, Lama and Akkaya, Ilge and Aleman, Florencia Leoni and Almeida, Diogo and Altenschmidt, Janko and Altman, Sam and Anadkat, Shyamal and others},
  journal={arXiv preprint arXiv:2303.08774},
  year={2023}
}

@article{touvron2023llama,
  title={Llama: Open and efficient foundation language models},
  author={Touvron, Hugo and Lavril, Thibaut and Izacard, Gautier and Martinet, Xavier and Lachaux, Marie-Anne and Lacroix, Timoth{\'e}e and Rozi{\`e}re, Baptiste and Goyal, Naman and Hambro, Eric and Azhar, Faisal and others},
  journal={arXiv preprint arXiv:2302.13971},
  year={2023}
}

@article{team2023gemini,
  title={Gemini: a family of highly capable multimodal models},
  author={Team, Gemini and Anil, Rohan and Borgeaud, Sebastian and Alayrac, Jean-Baptiste and Yu, Jiahui and Soricut, Radu and Schalkwyk, Johan and Dai, Andrew M and Hauth, Anja and Millican, Katie and others},
  journal={arXiv preprint arXiv:2312.11805},
  year={2023}
}

@article{comanici2025gemini,
  title={Gemini 2.5: Pushing the frontier with advanced reasoning, multimodality, long context, and next generation agentic capabilities},
  author={Comanici, Gheorghe and Bieber, Eric and Schaekermann, Mike and Pasupat, Ice and Sachdeva, Noveen and Dhillon, Inderjit and Blistein, Marcel and Ram, Ori and Zhang, Dan and Rosen, Evan and others},
  journal={arXiv preprint arXiv:2507.06261},
  year={2025}
}

@article{hu2021lora,
  title={Lora: Low-rank adaptation of large language models},
  author={Hu, Edward J and Shen, Yelong and Wallis, Phillip and Allen-Zhu, Zeyuan and Li, Yuanzhi and Wang, Shean and Wang, Lu and Chen, Weizhu},
  journal={arXiv preprint arXiv:2106.09685},
  year={2021}
}

@article{bai2025qwen2,
  title={Qwen2. 5-vl technical report},
  author={Bai, Shuai and Chen, Keqin and Liu, Xuejing and Wang, Jialin and Ge, Wenbin and Song, Sibo and Dang, Kai and Wang, Peng and Wang, Shijie and Tang, Jun and others},
  journal={arXiv preprint arXiv:2502.13923},
  year={2025}
}

@article{chen2024expanding,
  title={Expanding performance boundaries of open-source multimodal models with model, data, and test-time scaling},
  author={Chen, Zhe and Wang, Weiyun and Cao, Yue and Liu, Yangzhou and Gao, Zhangwei and Cui, Erfei and Zhu, Jinguo and Ye, Shenglong and Tian, Hao and Liu, Zhaoyang and others},
  journal={arXiv preprint arXiv:2412.05271},
  year={2024}
}

@inproceedings{camgoz2020sign,
  title={Sign language transformers: Joint end-to-end sign language recognition and translation},
  author={Camgoz, Necati Cihan and Koller, Oscar and Hadfield, Simon and Bowden, Richard},
  booktitle={CVPR},
  year={2020}
}

@article{lin2023gloss,
  title={Gloss-free end-to-end sign language translation},
  author={Lin, Kezhou and Wang, Xiaohan and Zhu, Linchao and Sun, Ke and Zhang, Bang and Yang, Yi},
  journal={arXiv preprint arXiv:2305.12876},
  year={2023}
}

@article{zhang2023sltunet,
  title={SLTUNET: A simple unified model for sign language translation},
  author={Zhang, Biao and M{\"u}ller, Mathias and Sennrich, Rico},
  journal={arXiv preprint arXiv:2305.01778},
  year={2023}
}

@inproceedings{Chen2022MMTLB,
  title={A simple multi-modality transfer learning baseline for sign language translation},
  author={Chen, Yutong and Wei, Fangyun and Sun, Xiao and Wu, Zhirong and Lin, Stephen},
  booktitle={CVPR},
  year={2022}
}

@article{li2025uni,
  title={Uni-sign: Toward unified sign language understanding at scale},
  author={Li, Zecheng and Zhou, Wengang and Zhao, Weichao and Wu, Kepeng and Hu, Hezhen and Li, Houqiang},
  journal={arXiv preprint arXiv:2501.15187},
  year={2025}
}

@inproceedings{zhou2023GFSLTVLP,
  title={Gloss-free sign language translation: Improving from visual-language pretraining},
  author={Zhou, Benjia and Chen, Zhigang and Clap{\'e}s, Albert and Wan, Jun and Liang, Yanyan and Escalera, Sergio and Lei, Zhen and Zhang, Du},
  booktitle={ICCV},
  year={2023}
}

@article{wong2024sign2gpt,
  title={Sign2GPT: Leveraging large language models for gloss-free sign language translation},
  author={Wong, Ryan and Camgoz, Necati Cihan and Bowden, Richard},
  journal={arXiv preprint arXiv:2405.04164},
  year={2024}
}

@inproceedings{duarte2021how2sign,
  title={How2sign: a large-scale multimodal dataset for continuous american sign language},
  author={Duarte, Amanda and Palaskar, Shruti and Ventura, Lucas and Ghadiyaram, Deepti and DeHaan, Kenneth and Metze, Florian and Torres, Jordi and Giro-i-Nieto, Xavier},
  booktitle={CVPR},
  year={2021}
}

@inproceedings{shi2022open,
  title={Open-domain sign language translation learned from online video},
  author={Shi, Bowen and Brentari, Diane and Shakhnarovich, Gregory and Livescu, Karen},
  booktitle={EMNLP},
  year={2022}
}

@article{rust2024towards,
  title={Towards privacy-aware sign language translation at scale},
  author={Rust, Phillip and Shi, Bowen and Wang, Skyler and Camg{\"o}z, Necati Cihan and Maillard, Jean},
  journal={arXiv preprint arXiv:2402.09611},
  year={2024}
}

@inproceedings{gueuwou2025shubert,
  title={SHuBERT: Self-supervised sign language representation learning via multi-stream cluster prediction},
  author={Gueuwou, Shester and Du, Xiaodan and Shakhnarovich, Greg and Livescu, Karen and Liu, Alexander H},
  booktitle={ACL},
  year={2025}
}

@article{liu2023visual,
  title={Visual instruction tuning},
  author={Liu, Haotian and Li, Chunyuan and Wu, Qingyang and Lee, Yong Jae},
  journal={Neurips},
  volume={36},
  year={2023}
}

@article{li2024llava,
  title={Llava-onevision: Easy visual task transfer},
  author={Li, Bo and Zhang, Yuanhan and Guo, Dong and Zhang, Renrui and Li, Feng and Zhang, Hao and Zhang, Kaichen and Zhang, Peiyuan and Li, Yanwei and Liu, Ziwei and others},
  journal={arXiv preprint arXiv:2408.03326},
  year={2024}
}

@article{bai2023qwen,
  title={Qwen technical report},
  author={Bai, Jinze and Bai, Shuai and Chu, Yunfei and Cui, Zeyu and Dang, Kai and Deng, Xiaodong and Fan, Yang and Ge, Wenbin and Han, Yu and Huang, Fei and others},
  journal={arXiv preprint arXiv:2309.16609},
  year={2023}
}

@inproceedings{kim2025leveraging,
  title={Leveraging the power of mllms for gloss-free sign language translation},
  author={Kim, Jungeun and Jeon, Hyeongwoo and Bae, Jongseong and Kim, Ha Young},
  booktitle={ICCV},
  year={2025}
}

@inproceedings{papineni2002bleu,
  title={Bleu: a method for automatic evaluation of machine translation},
  author={Papineni, Kishore and Roukos, Salim and Ward, Todd and Zhu, Wei-Jing},
  booktitle={ACL},
  year={2002}
}

@inproceedings{lin2004rouge,
  title={Rouge: A package for automatic evaluation of summaries},
  author={Lin, Chin-Yew},
  booktitle={Text summarization branches out},
  pages={74--81},
  year={2004}
}

@inproceedings{sellam2020bleurt,
  title={BLEURT: Learning robust metrics for text generation},
  author={Sellam, Thibault and Das, Dipanjan and Parikh, Ankur},
  booktitle={ACL},
  year={2020}
}

@inproceedings{tarres2023sign,
  title={Sign language translation from instructional videos},
  author={Tarr{\'e}s, Laia and G{\'a}llego, Gerard I and Duarte, Amanda and Torres, Jordi and Gir{\'o}-i-Nieto, Xavier},
  booktitle={CVPR},
  year={2023}
}

@article{uthus2023youtube,
  title={Youtube-asl: A large-scale, open-domain american sign language-english parallel corpus},
  author={Uthus, Dave and Tanzer, Garrett and Georg, Manfred},
  journal={Neurips},
  volume={36},
  year={2023}
}

@inproceedings{gueuwou2025signmusketeers,
  title={SignMusketeers: An efficient multi-stream approach for sign language translation at scale},
  author={Gueuwou, Shester and Du, Xiaodan and Shakhnarovich, Greg and Livescu, Karen},
  booktitle={ACL},
  year={2025}
}

@article{chen2025c,
  title={C 2 RL: Content and Context Representation Learning for Gloss-free Sign Language Translation and Retrieval},
  author={Chen, Zhigang and Zhou, Benjia and Huang, Yiqing and Wan, Jun and Hu, Yibo and Shi, Hailin and Liang, Yanyan and Lei, Zhen and Zhang, Du},
  journal={IEEE Transactions on Circuits and Systems for Video Technology},
  year={2025},
  publisher={IEEE}
}

@article{zhang2024scaling,
  title={Scaling sign language translation},
  author={Zhang, Biao and Tanzer, Garrett and Firat, Orhan},
  journal={Neurips},
  volume={37},
  year={2024}
}

@inproceedings{gong2024llms,
  title={Llms are good sign language translators},
  author={Gong, Jia and Foo, Lin Geng and He, Yixuan and Rahmani, Hossein and Liu, Jun},
  booktitle={CVPR},
  year={2024}
}

@inproceedings{chen2024internvl,
  title={Internvl: Scaling up vision foundation models and aligning for generic visual-linguistic tasks},
  author={Chen, Zhe and Wu, Jiannan and Wang, Wenhai and Su, Weijie and Chen, Guo and Xing, Sen and Zhong, Muyan and Zhang, Qinglong and Zhu, Xizhou and Lu, Lewei and others},
  booktitle={CVPR},
  year={2024}
}

@article{shao2024deepseekmath,
  title={Deepseekmath: Pushing the limits of mathematical reasoning in open language models},
  author={Shao, Zhihong and Wang, Peiyi and Zhu, Qihao and Xu, Runxin and Song, Junxiao and Bi, Xiao and Zhang, Haowei and Zhang, Mingchuan and Li, YK and Wu, Yang and others},
  journal={arXiv preprint arXiv:2402.03300},
  year={2024}
}

@inproceedings{gan2023realtime,
  title={Towards Real-Time Sign Language Recognition and Translation on Edge Devices},
  author={Gan, Shiwei and Yin, Yafeng and Jiang, Zhiwei and Xie, Lei and Lu, Sanglu},
  booktitle={Proceedings of the 31st ACM International Conference on Multimedia},
  pages={4502--4512},
  year={2023}
}

@article{liddell1989american,
  title={American Sign Language: The Phonological Base},
  author={Liddell, Scott K. and Johnson, Robert E.},
  journal={Sign Language Studies},
  volume={64},
  number={1},
  pages={195--277},
  year={1989}
}

@article{brookshire2017visual,
  title={Visual cortex entrains to sign language},
  author={Brookshire, Geoffrey and Lu, Jenny and Nusbaum, Howard C. and Goldin-Meadow, Susan and Casasanto, Daniel},
  journal={Proceedings of the National Academy of Sciences},
  volume={114},
  number={24},
  pages={6352--6357},
  year={2017},
  doi={10.1073/pnas.1620350114}
}

@article{yu2025dapo,
  title={DAPO: An Open-Source LLM Reinforcement Learning System at Scale},
  author={Yu, Qiying and Zhang, Zheng and Zhu, Ruofei and Yuan, Yufeng and Zuo, Xiaochen and Yue, Yu and Fan, Tiantian and Liu, Gaohong and Liu, Lingjun and Liu, Xin and Lin, Haibin and Lin, Zhiqi and Ma, Xuechi and Sheng, Dawei and Tong, Yuxuan and Yuan, Chi and Zhou, Hao and Wang, Minging and Li, Weinan and Li, Zhi and Liu, Heung-Yeung and Wang, Mingsheng and Chen, Weizhu and Lin, Yu},
  journal={arXiv preprint arXiv:2503.14476},
  year={2025}
}

@article{hu2025openreasonerzero,
  title={Open-Reasoner-Zero: An Open Source Approach to Scaling Up Reinforcement Learning on the Base Model},
  author={Hu, Jingcheng and Zhang, Yinmin and Han, Qi and Jiang, Daxin and Zhang, Xiangyu and Shum, Heung-Yeung},
  journal={arXiv preprint arXiv:2503.24290},
  year={2025}
}

@inproceedings{post2018call,
  title={A call for clarity in reporting BLEU scores},
  author={Post, Matt},
  booktitle={Proceedings of the third conference on machine translation: Research papers},
  pages={186--191},
  year={2018}
}

@article{hu2023signbert+,
  title={Signbert+: Hand-model-aware self-supervised pre-training for sign language understanding},
  author={Hu, Hezhen and Zhao, Weichao and Zhou, Wengang and Li, Houqiang},
  journal={IEEE Transactions on Pattern Analysis and Machine Intelligence},
  volume={45},
  number={9},
  pages={11221--11239},
  year={2023},
  publisher={IEEE}
}

@article{rao2025rvlf,
  title={RVLF: A Reinforcing Vision-Language Framework for Gloss-Free Sign Language Translation},
  author={Rao, Zhi and Zhou, Yucheng and Zhou, Benjia and Huang, Yiqing and Escalera, Sergio and Wan, Jun},
  journal={arXiv preprint arXiv:2512.07273},
  year={2025}
}

@inproceedings{selvaraju2019hint,
  title={Taking a hint: Leveraging explanations to make vision and language models more grounded},
  author={Selvaraju, Ramprasaath R and Lee, Stefan and Shen, Yilin and Jin, Hongxia and Ghosh, Shalini and Heck, Larry and Batra, Dhruv and Parikh, Devi},
  booktitle={Proceedings of the IEEE/CVF international conference on computer vision},
  pages={2591--2600},
  year={2019}
}

@inproceedings{wan2024contrastive,
  title={Contrastive region guidance: Improving grounding in vision-language models without training},
  author={Wan, David and Cho, Jaemin and Stengel-Eskin, Elias and Bansal, Mohit},
  booktitle={European Conference on Computer Vision},
  pages={198--215},
  year={2024},
  organization={Springer}
}

@article{li2024surprising,
  title={On the surprising effectiveness of attention transfer for vision transformers},
  author={Li, Alexander C and Tian, Yuandong and Chen, Beidi and Pathak, Deepak and Chen, Xinlei},
  journal={Advances in Neural Information Processing Systems},
  volume={37},
  pages={113963--113990},
  year={2024}
}

@article{yan2024voila,
  title={Voila-a: Aligning vision-language models with user's gaze attention},
  author={Yan, Kun and Wang, Zeyu and Ji, Lei and Wang, Yuntao and Duan, Nan and Ma, Shuai},
  journal={Advances in neural information processing systems},
  volume={37},
  pages={1890--1918},
  year={2024}
}

@book{brentari2010handshape,
  title={Handshape contrasts in sign language phonology},
  author={Brentari, Diane and Eccarius, Petra},
  year={2010},
  publisher={na}
}

@book{pfau2010nonmanuals,
  title={Nonmanuals: Their grammatical and prosodic roles},
  author={Pfau, Roland and Quer, Josep and others},
  year={2010},
  publisher={na}
}

@article{guo2026bridging,
  title={Bridging sign and spoken languages: Pseudo gloss generation for sign language translation},
  author={Guo, Jianyuan and Li, Peike and Cohn, Trevor},
  journal={Advances in Neural Information Processing Systems},
  volume={38},
  pages={77471--77499},
  year={2026}
}

@article{fish2026geo,
  title={Geo-sign: Hyperbolic contrastive regularisation for geometrically aware sign language translation},
  author={Fish, Edward and Bowden, Richard},
  journal={Advances in Neural Information Processing Systems},
  volume={38},
  pages={99293--99330},
  year={2026}
}

@article{pu2026signdpo,
  title={SignDPO: Multi-level Direct Preference Optimisation for Skeleton-based Gloss-free Sign Language Translation},
  author={Pu, Muxin and Wu, Xiao-Ming and Lim, Mei Kuan and Chong, Chun Yong and Li, Wei and Loy, Chen Change},
  journal={arXiv preprint arXiv:2604.18034},
  year={2026}
}

@article{hwang2024efficient,
  title={An efficient sign language translation using spatial configuration and motion dynamics with llms},
  author={Hwang, Eui Jun and Cho, Sukmin and Lee, Junmyeong and Park, Jong C},
  journal={arXiv preprint arXiv:2408.10593},
  year={2024}
}

@article{lugaresi2019mediapipe,
  title={Mediapipe: A framework for building perception pipelines},
  author={Lugaresi, Camillo and Tang, Jiuqiang and Nash, Hadon and McClanahan, Chris and Uboweja, Esha and Hays, Michael and Zhang, Fan and Chang, Chuo-Ling and Yong, Ming Guang and Lee, Juhyun and others},
  journal={arXiv preprint arXiv:1906.08172},
  year={2019}
}

@article{oquab2023dinov2,
  title={Dinov2: Learning robust visual features without supervision},
  author={Oquab, Maxime and Darcet, Timoth{\'e}e and Moutakanni, Th{\'e}o and Vo, Huy and Szafraniec, Marc and Khalidov, Vasil and Fernandez, Pierre and Haziza, Daniel and Massa, Francisco and El-Nouby, Alaaeldin and others},
  journal={arXiv preprint arXiv:2304.07193},
  year={2023}
}

@article{liu2020multilingual,
  title={Multilingual denoising pre-training for neural machine translation},
  author={Liu, Yinhan and Gu, Jiatao and Goyal, Naman and Li, Xian and Edunov, Sergey and Ghazvininejad, Marjan and Lewis, Mike and Zettlemoyer, Luke},
  journal={Transactions of the Association for Computational Linguistics},
  volume={8},
  pages={726--742},
  year={2020},
  publisher={MIT Press One Rogers Street, Cambridge, MA 02142-1209, USA journals-info~…}
}

@inproceedings{he2016deep,
  title={Deep residual learning for image recognition},
  author={He, Kaiming and Zhang, Xiangyu and Ren, Shaoqing and Sun, Jian},
  booktitle={Proceedings of the IEEE conference on computer vision and pattern recognition},
  pages={770--778},
  year={2016}
}

@article{liang2024llava,
  title={Llava-slt: Visual language tuning for sign language translation},
  author={Liang, Han and Huang, Chengyu and Xu, Yuecheng and Tang, Cheng and Ye, Weicai and Zhang, Juze and Chen, Xin and Yu, Jingyi and Xu, Lan},
  journal={arXiv preprint arXiv:2412.16524},
  year={2024}
}

@article{feng2026video,
  title={Video-r1: Reinforcing video reasoning in mllms},
  author={Feng, Kaituo and Gong, Kaixiong and Li, Bohao and Guo, Zonghao and Wang, Yibing and Peng, Tianshuo and Wu, Junfei and Zhang, Xiaoying and Wang, Benyou and Yue, Xiangyu},
  journal={Advances in Neural Information Processing Systems},
  volume={38},
  pages={99114--99137},
  year={2026}
}

@inproceedings{jang2025lost,
  title={Lost in translation, found in context: Sign language translation with contextual cues},
  author={Jang, Youngjoon and Raajesh, Haran and Momeni, Liliane and Varol, G{\"u}l and Zisserman, Andrew},
  booktitle={Proceedings of the Computer Vision and Pattern Recognition Conference},
  pages={8742--8752},
  year={2025}
}

@inproceedings{hamidullah2024sign,
  title={Sign language translation with sentence embedding supervision},
  author={Hamidullah, Yasser and van Genabith, Josef and Espa{\~n}a-Bonet, Cristina},
  booktitle={Proceedings of the 62nd Annual Meeting of the Association for Computational Linguistics (Volume 2: Short Papers)},
  pages={425--434},
  year={2024}
}

@article{karakucs2026semantic,
  title={Semantic Communities and Boundary-Spanning Lyrics in K-pop: A Graph-Based Unsupervised Analysis},
  author={Karaku{\c{s}}, Oktay},
  journal={arXiv preprint arXiv:2602.12881},
  year={2026}
}

@article{jiang2026think,
  title={Think in Latent Thoughts: A New Paradigm for Gloss-Free Sign Language Translation},
  author={Jiang, Yiyang and Zhang, Li and Wei, Xiao-Yong and Qing, Li},
  journal={arXiv preprint arXiv:2604.15301},
  year={2026}
}

@inproceedings{zhao2026cnsl,
  title={CNSL-bench: Benchmarking the Sign Language Understanding Capabilities of MLLMs on Chinese National Sign Language},
  author={Zhao, Rui and Zhong, Xuewen and Zheng, Xiaoyun and Su, Jinsong and Chen, Yidong},
  booktitle={Proceedings of the 64th Annual Meeting of the Association for Computational Linguistics (Volume 1: Long Papers)},
  year={2026}
}

@inproceedings{inan2025signalignlm,
  title={SignAlignLM: Integrating multimodal sign language processing into large language models},
  author={Inan, Mert and Sicilia, Anthony and Alikhani, Malihe},
  booktitle={Findings of the Association for Computational Linguistics: ACL 2025},
  pages={3691--3706},
  year={2025}
}

\clearpage
\appendix

\section{Training Instructions}
\label{app:instructions}

We use diverse instruction phrasings during instruction tuning to improve robustness to prompt variation. Examples include:

\begin{enumerate}[noitemsep, topsep=2pt, leftmargin=*, label=(\arabic*)]
    \item ``Translate the American Sign Language in this video to English.''
    \item ``What is being signed in this video? Provide the English translation.''
    \item ``Watch the sign language video and write the English sentence.''
\end{enumerate}

\section{Further Discussion on the Vision Encoder}
\label{app:vit_freezing}

\begin{table}[h]
\centering
\footnotesize
\setlength{\tabcolsep}{5pt}
\begin{tabular}{lcccc}
\toprule
ViT Setting & B-1 & B-4 & ROUGE-L & BLEURT \\
\midrule
Freeze              & 20.45 & 1.79 & 16.89 & 30.11 \\
Half (top-6)        & 22.85 & 3.87 & 18.97 & 31.20 \\
\rowcolor{gray!15}
Full (Ours)         & \textbf{32.61} & \textbf{8.06} & \textbf{28.47} & \textbf{36.82} \\
\bottomrule
\end{tabular}
\caption{ViT freezing strategy on H2S. Fully tuning the ViT is critical for adapting to the sign-language visual distribution.}
\label{tab:vit}
\end{table}

During the SFT stage, we study how much of the ViT should be trained in our task, comparing three strategies: keeping the entire ViT frozen, unfreezing only its top six transformer layers, and fully fine-tuning all ViT parameters jointly with the LLM. All variants share identical training data and hyperparameters; only the ViT trainable parameters differ.
Table~\ref{tab:vit} shows that full fine-tuning of the ViT is essential for SLT. A fully frozen ViT yields only 1.79 BLEU-4, indicating that the pretrained general-image features are far from sufficient for the fine-grained hand and facial cues in signing videos. Partially unfreezing the top six layers improves BLEU-4 to 3.87 but still falls short of the fully fine-tuned setting with BLEU-4 8.06. We therefore fully fine-tune the ViT alongside the LLM throughout all of our experiments.

\section{Further Discussion on Spatial Attention Loss}
\label{app:kl_vs_bg}

\begin{table}[h]
\centering
\small
\setlength{\tabcolsep}{4pt}
\resizebox{\columnwidth}{!}{%
\begin{tabular}{lcccc}
\toprule
Attention loss & B-1 & B-4 & ROUGE-L & BLEURT \\
\midrule
SFT (baseline)              & 32.61 & 8.06 & 28.47 & 36.82 \\
\;+ Background Penalty      & 32.35 & 7.81 & 27.42 & 36.47 \\
\rowcolor{gray!15}
\;+ KL Loss (Ours)          & \textbf{34.16} & \textbf{8.69} & \textbf{29.31} & \textbf{42.21} \\
\bottomrule
\end{tabular}
}
\caption{Comparison of two attention supervision losses.}
\label{tab:kl_vs_bg}
\end{table}

In the SA-SFT stage, our spatial attention supervision uses a KL loss via a soft foreground target. An alternative is a direct background penalty that directly suppresses attention mass outside the foreground mask:
\begin{equation}
    \mathcal{L}_{\mathrm{bg}} = \sum_{i \notin \mathrm{FG}} P^{(l)}_{\mathrm{attn},i}.
\end{equation}

The background penalty has a degenerate solution: the loss can be minimized by placing all attention on few foreground tokens, which discards information from all other foreground regions. We observe this collapse at the supervised layers, which end up receiving little visual signal. As a result, the background penalty underperforms even plain SFT, dropping BLEU-4 from 8.06 to 7.81. In contrast, the KL loss matches the entire attention distribution against the soft target $\hat{P}_k$, which prevents collapse onto a single token and keeps attention distributed across the hands and face, yielding better performance.

\section{Foreground Mask Generation Details}
\label{app:mask_details}

We use MediaPipe~\cite{lugaresi2019mediapipe} to construct per-frame foreground masks for both How2Sign and OpenASL training sets. MediaPipe has been widely adopted in recent SLT works~\cite{gueuwou2025shubert,gueuwou2025signmusketeers}. MediaPipe's FaceDetection and Hands modules are widely used for their accurate, real-time detection of face and hand regions in videos, with reliable performance across diverse signers and camera setups. For each frame, we extract bounding boxes for the face and both hands and build a binary mask at the original frame resolution by marking all pixels inside these boxes as foreground. The mask is then average-pooled to the $8\times8$ visual-token grid and stored on disk before training. When MediaPipe fails to produce a valid mask for a video, we skip the spatial attention loss for that sample and retain only the cross-entropy term.

\section{Further Discussion on Softmax Temperature}
\label{app:T_sweep}

\begin{figure}[h]
\centering
\includegraphics[width=0.75\columnwidth]{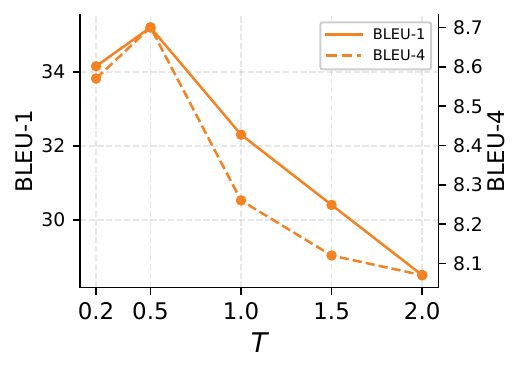}
\vspace{-3mm}
\caption{Effect of varying the softmax temperature $T$ in the spatial attention target on H2S during SA-SFT, with $\gamma{=}0.15$ fixed.}
\label{fig:T_sweep}
\end{figure}

In the SA-SFT stage, the softmax temperature $T$ controls how sharply the spatial target $\hat{P}_k$ concentrates on foreground tokens: smaller $T$ produces a sharper target focused on the hand and face regions, while larger $T$ flattens the target toward a uniform distribution over all tokens. We sweep different temperature values on H2S, with the loss weight fixed at $\gamma{=}0.15$ (Figure~\ref{fig:T_sweep}). $T{=}0.5$ achieves the best BLEU-4 of 8.69, with $T{=}0.2$ close behind, indicating that a relatively sharp foreground target is effective. As $T$ increases beyond 0.7, performance drops back toward the baseline level (8.06): the target becomes too flat to provide useful guidance and the attention supervision becomes ineffective. We therefore use $T{=}0.5$ in all main experiments.

\section{Spatial Attention Visualization on OpenASL}
\label{app:openasl_attn}

\begin{figure}[h]
    \centering
    \includegraphics[width=\columnwidth]{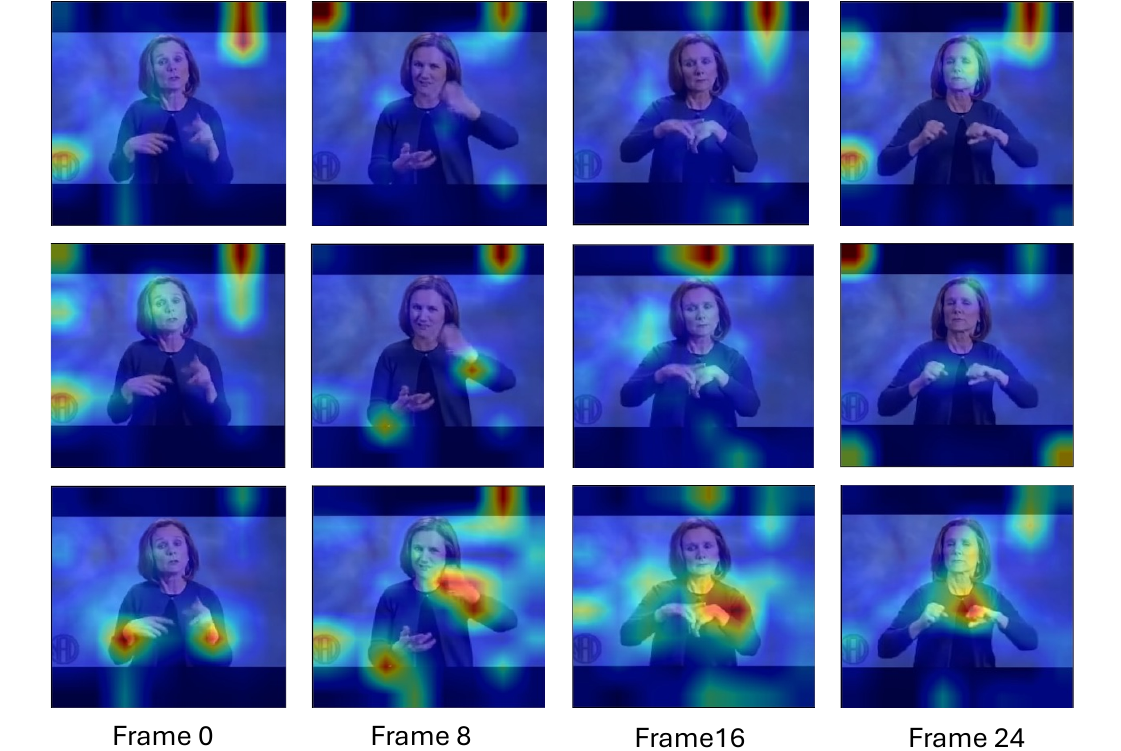}
    \caption{Visual-token attention maps at LLM layer 16 across four frames (columns) of an OpenASL test video. Rows show the three training stages: pre-trained VLM (top), standard SFT (middle), and Sign-Aware SFT with our spatial attention supervision (bottom).}
    \label{fig:openasl_attn}
    \vspace{-3mm}
\end{figure}

To verify that the spatial attention supervision generalizes beyond H2S, we visualize attention maps on an OpenASL test video. Figure~\ref{fig:openasl_attn} mirrors the layout of Figure~\ref{fig:attn_heatmap_analysis}. As on H2S, SA-SFT concentrates attention on the signer's hands and face, while the pre-trained VLM and standard SFT model fail to focus, indicating that the spatial steering effect generalizes to OpenASL.

\section{Further Discussion on RL Training Data}
\label{app:grpo_sweet}

\begin{figure}[h]
\centering
\includegraphics[width=0.67\columnwidth]{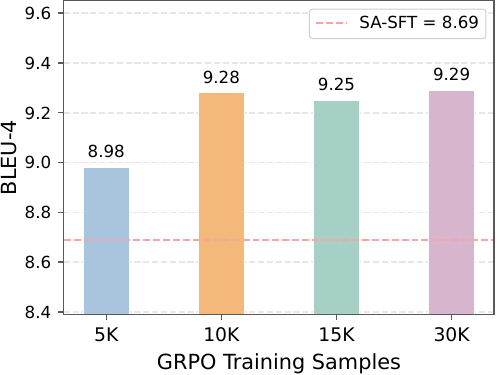}
\vspace{-2mm}
\caption{Effect of GRPO training set size on H2S, evaluated under standard GRPO. Performance plateaus beyond 10K samples.}
\label{fig:grpo_size}
\end{figure}

In the RL stage, we sweep the number of in-domain training samples across 5K, 10K, 15K, and 30K under standard GRPO (without our motion-cadence steering), to isolate the effect of data size from other design choices. Figure~\ref{fig:grpo_size} shows that performance improves from 5K to 10K and then plateaus, with 15K and 30K offering no additional gain over 10K. We therefore use 10K samples as our default GRPO setting; additional samples beyond this size yield no improvement and only increase training cost.

\section{Motion-Cadence Target Ablation}
\label{app:peak_source}

In the SA-GRPO stage, the motion-cadence target $t_k$ is built from detected motion peaks plus Gaussian smoothing. To check whether the gain comes from the content-aligned peak positions, we replace the detected peaks with random peak positions and keep the rest of the cadence loss (KL form, Gaussian smoothing, peak count, weight $\lambda$) unchanged.

\begin{table}[h]
\centering

\small
\setlength{\tabcolsep}{4pt}
\resizebox{\columnwidth}{!}{%
\begin{tabular}{lcccc}
\toprule
Target source & B-1 & B-4 & ROUGE-L & BLEURT \\
\midrule
GRPO w/o KL & 35.62 & 9.28 & 30.07 & 42.62 \\
\midrule
random peaks  & 35.59    & 9.21    & 29.84    & 42.59    \\
\rowcolor{gray!15}
motion peaks (Ours) & \textbf{35.82} & \textbf{9.45} & \textbf{30.45} & \textbf{42.73} \\
\bottomrule
\end{tabular}
}
\caption{Effect of the motion-cadence target source on H2S. Random peaks share the same number and Gaussian smoothing as the detected peaks; only the alignment with video content is removed.}
\label{tab:peak_source}
\end{table}

Random-peak target performs on par with the GRPO w/o KL baseline (BLEU-4 $9.21$ vs.\ $9.28$), indicating that adding a cadence loss with an arbitrary target brings no benefit on its own. This shows that the gain comes from the content-aligned peak positions specifically, not from adding any attention regularization in the RL stage.

\section{Further Discussion on Supervised Layers for SA-GRPO}
\label{app:layer_falsify}

In the SA-GRPO stage, the statistical analysis selects the 17th and 19th layers based on their positive concentration-quality correlation. To check that the probe genuinely predicts the best layers, we move the motion-cadence steering to the 5th and 7th layers, where the analysis shows weak or negative correlation.

\begin{table}[h]
\centering
\footnotesize
\setlength{\tabcolsep}{5pt}
\begin{tabular}{lcccc}
\toprule
Layer & B-1 & B-4 & ROUGE-L & BLEURT \\
\midrule
5/7           & 35.52 & 9.19 & 29.69 & 42.51 \\
\rowcolor{gray!15}
17/19 (Ours)  & \textbf{35.82} & \textbf{9.45} & \textbf{30.45} & \textbf{42.73} \\
\bottomrule
\end{tabular}
\caption{Layer-selection falsification for SA-GRPO.}
\label{tab:layer_falsify}
\end{table}

SA-GRPO at the 5th and 7th layers falls below the 17th and 19th layers across all metrics, consistent with the probe's prediction. This confirms the probe is a predictive signal for layer selection.

\section{Ablation Study on OpenASL}
\label{app:openasl_ablation}

To verify that the component-wise gains observed on H2S are not specific to that dataset, we run the same ablation on OpenASL. SFT is trained on OpenASL and YT, GRPO is trained on 10K OpenASL in-domain samples, and evaluation is on the OpenASL test set. We reuse the same supervised layers as on H2S to test cross-dataset transferability.

\begin{table}[h]
\centering
\small
\setlength{\tabcolsep}{4pt}
\resizebox{\columnwidth}{!}{%
\begin{tabular}{lcccc}
\toprule
Method & B-1 & B-4 & ROUGE-L & BLEURT \\
\midrule
Zero-shot baseline  & 11.23    & 0.35    & 5.67  & 19.76\\
\midrule
\multicolumn{5}{l}{\textit{SFT stage}} \\
\quad SFT    & 45.67    & 18.92    &  41.85    & 58.76    \\
\rowcolor{gray!15}
\quad SA-SFT    & 46.89    & 19.76    & 43.53 &  60.23    \\
\midrule
\multicolumn{5}{l}{\textit{GRPO stage}} \\
\quad GRPO (w/ KL)   & 47.03    & 20.04    & 44.75     & 60.98     \\
\quad GRPO (w/o KL)  &  47.51 & 20.55  & 45.34   & 61.56     \\
\quad GRPO (w/ KL + cadence)   & 47.12 & 20.22  & 45.26     & 61.31     \\
\rowcolor{gray!15}
\quad SA-GRPO (Ours) & \textbf{47.74} & \textbf{20.75} & \textbf{45.62} & \textbf{61.68} \\
\bottomrule
\end{tabular}
}

\caption{Component ablation on OpenASL. The pattern of gains mirrors that on H2S (Table~\ref{tab:component_ablation}): SA-SFT improves over plain SFT, removing the KL penalty further improves over standard GRPO, and SA-GRPO yields the best results across all metrics.}
\label{tab:openasl_ablation}
\end{table}

The component-wise gain pattern on OpenASL matches that on How2Sign, confirming that the contributions of SA-SFT and SA-GRPO generalize across both benchmarks rather than reflecting How2Sign-specific tuning.

\section{Qualitative Results}
\label{app:translations}

We show qualitative translation examples on How2Sign (Table~\ref{tab:examples_h2s}) and OpenASL (Table~\ref{tab:examples_openasl}), with each example reporting the ground-truth reference alongside the output of AttnSign. AttnSign successfully captures the semantic information in most examples and generates sentences close to the ground-truth references, while still struggling on more complex cases, as shown in the last example of each table.

\begin{table}[h]
\centering
\small
\setlength{\tabcolsep}{4pt}
\renewcommand{\arraystretch}{1.15}
\begin{tabular}{l p{6.0cm}}
\toprule
Reference: & I'm CarolAnn with Studio Group X and just do those push ups every day and you'll improve. \\
AttnSign:  & I'm Carol Ann with Studio Group X and we just did those pushups every day and you will improve. \\
\midrule
Reference: & It has a bright color due to the fruit punch. \\
AttnSign:  & It has a bright color because of the pineapple. \\
\midrule
Reference: & I practice with the Barton Oaks Dental Group. \\
AttnSign:  & I practice with the Barton Oaks Dental Groups. \\
\midrule
Reference: & To come out of it, soften the knees, take the hands away from your feet, and slowly roll it up, tuck in the tail bone and rolling up one vertebra at a time and come straight up. \\
AttnSign:  & And then she comes out of the tension soft hands to the floor and slowly roll her knee and she loosens her bones as she rolls up into a lunge. \\
\bottomrule
\end{tabular}
\caption{Qualitative results on the How2Sign dataset.}
\label{tab:examples_h2s}
\end{table}

\begin{table}[h]
\centering
\small
\setlength{\tabcolsep}{4pt}
\renewcommand{\arraystretch}{1.15}
\begin{tabular}{l p{6.0cm}}
\toprule
Reference: & We have a lot of respect for our Canadian neighbors. \\
AttnSign:  & We have a lot of respect for our neighbors. \\
\midrule
Reference: & His family were skilled woodcarvers and sold totems around Seattle since 1909, passing down their carving skills. \\
AttnSign:  & His family was skilled woodcarvers and sold to totem poles around Seattle area since 1909, as an art skills. \\
\midrule
Reference: & Biden had 150,000 more votes than Bernie. \\
AttnSign:  & Biden is ahead by 150,000 votes than Bernie. \\
\midrule
Reference: & You need to invest your time and passion to assist delegates to understand why this topic is important to ensure your proposed priority successfully ranks. \\
AttnSign:  & We want you to spend your time and make sure you understand your passion, your desires, and your work to be successful in proposing an idea and getting it passed. \\
\bottomrule
\end{tabular}
\caption{Qualitative results on the OpenASL dataset.}
\label{tab:examples_openasl}
\end{table}

\end{document}